\documentclass[11pt,a4paper]{article}
\usepackage[hyperref]{acl2019}
\usepackage{times}
\usepackage{latexsym}

\usepackage{url}

\usepackage{amsmath}
\usepackage{amssymb}
\usepackage{multirow}
\usepackage{array}
\usepackage{graphicx}
\usepackage{comment}
\usepackage{booktabs}
\usepackage{subfigure}
\usepackage{caption}
\usepackage{xcolor,colortbl} 
\usepackage{multirow}
\usepackage{makecell}
\usepackage{enumitem}
\usepackage{resizegather}

\DeclareMathOperator*{\argmax}{arg\,max}

\aclfinalcopy 
\def\aclpaperid{1042} 

\title{Interconnected Question Generation with Coreference Alignment and Conversation Flow Modeling}
  
\author{Yifan Gao\textsuperscript{1}\thanks{~~~This work was partially done when Yifan Gao was an intern at Tencent AI Lab.} ~~~~~~ Piji Li\textsuperscript{2}
~~~~~~ Irwin King\textsuperscript{1} ~~~~~~ Michael R. Lyu\textsuperscript{1}  \\
	{\textsuperscript{1} Department of Computer Science and Engineering, }\\
	{The Chinese University of Hong Kong, Shatin, N.T., Hong Kong}\\
	{\textsuperscript{2} Tencent AI Lab}\\
    { \textsuperscript{1}\{yfgao,king,lyu\}@cse.cuhk.edu.hk}
	{ \textsuperscript{2}pijili@tencent.com}
}

\date{}

\begin{document}
\maketitle
\begin{abstract}
We study the problem of generating interconnected questions in question-answering style conversations.
Compared with previous works which generate questions based on a single sentence (or paragraph), this setting is different in two major aspects: 
(1) Questions are highly conversational. Almost half of them refer back to conversation history using coreferences.
(2) In a coherent conversation, questions have smooth transitions between turns.
We propose an end-to-end neural model with coreference alignment and conversation flow modeling.
The coreference alignment modeling explicitly aligns coreferent mentions in conversation history with corresponding pronominal references in generated questions, which makes generated questions interconnected to conversation history.
The conversation flow modeling builds a coherent conversation by starting questioning on the first few sentences in a text passage and smoothly shifting the focus to later parts.
Extensive experiments show that our system outperforms several baselines and can generate highly conversational questions.
The code implementation is released at \url{https://github.com/Evan-Gao/conversational-QG}.
\end{abstract}

\section{Introduction}\label{sec.intro}

Question Generation (QG) aims to create human-like questions from a range of inputs, such as natural language text \cite{Heilman2010GoodQS}, knowledge base \cite{Serban2016GeneratingFQ} and image \cite{Mostafazadeh2016GeneratingNQ}.
QG is helpful for the knowledge testing in education, i.e., the intelligence tutor system, where an instructor can actively ask questions to students given reading comprehension materials \cite{Heilman2010GoodQS, Du2017LearningTA}. 
Besides, raising good questions in a conversational can enhance the interactiveness and persistence of human-machine interactions \cite{Wang2018LearningTA}.

\begin{table}
\centering
\scriptsize
\scalebox{0.93}{
\begin{tabular}{ll} 
\hline\hline
\multicolumn{2}{p{1.0\columnwidth}}{Passage: Incumbent Democratic President Bill Clinton was ineligible to serve a third term due to term limitations in the 22nd Amendment of the Constitution, and Vice President Gore was able to secure the Democratic nomination with relative ease. Bush was seen as the early favorite for the Republican nomination and, despite a contentious primary battle with Senator John McCain and other candidates, secured the nomination by Super Tuesday. Bush chose ...} \\
\hline
$\text{Q}_1$:~~ What political party is Clinton a member of? & $\text{A}_1$:~~ Democratic\\
$\text{Q}_2$:~~ What was he ineligible to serve?             & $\text{A}_2$:~~ third term\\
$\text{Q}_3$:~~ Why?                                         & $\text{A}_3$:~~ term limitations\\
$\text{Q}_4$:~~ Based on what amendment?                     & $\text{A}_4$:~~ 22nd\\
$\text{Q}_5$:~~ Of what document?                            & $\text{A}_5$:~~ Constitution\\
$\text{Q}_6$:~~ Who was his vice president?                  & $\text{A}_6$:~~ Gore\\
$\text{Q}_7$:~~ Who was the early Republican favorite for    & $\text{A}_7$:~~ Bush\\ 
~~~~~~~~~~the nomination?                              & \\
$\text{Q}_8$:~~ Who was the primary battle with?             & $\text{A}_8$:~~ John McCain\\
$\text{Q}_9$:~~ What is his title?                           & $\text{A}_9$:~~ Senator\\
$\text{Q}_{10}$: When did Bush secure the nomination by?       & $\text{A}_{10}$: Tuesday\\
\hline\hline
\end{tabular}
}
\caption{An example for conversational question generation from a conversational question answering dataset CoQA \cite{reddy2019coqa}. Each turn contains a question $\text{Q}_i$ and an answer $\text{A}_i$.}
\label{tab:example}
\end{table}

Recent works on question generation for knowledge testing are mostly formalized as a standalone interaction \cite{Yuan2017MachineCB,Song2018LeveragingCI}, while it is a more natural way for human beings to test knowledge or seek information through conversations involving a series of interconnected questions \cite{reddy2019coqa}. 
Furthermore, the inability for virtual assistants to ask questions based on previous discussions often leads to unsatisfying user experiences.
In this paper, we consider a new setting called \textbf{C}onversational \textbf{Q}uestion \textbf{G}eneration (\textbf{CQG}).
In this scenario, a system needs to ask a series of interconnected questions grounded in a passage through a question-answering style conversation. 
Table \ref{tab:example} provides an example under this scenario.
In this dialogue, a questioner and an answerer chat about the above passage. Every question after the first turn is dependent on the conversation history.

\begin{figure}[t!]
\centering
\includegraphics[width=1.0\columnwidth]{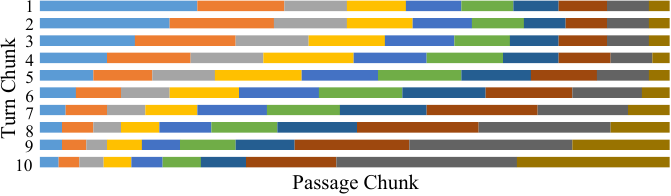}
\caption{
Passage chunks of interest for each turn chunks.
Each row contains 10 bands distinguished by different colors.
Each band represents a passage chunk.
The width of a passage chunk indicates the concentration of conversation in that turn.
The $y$-axis indicates turn chunk number.
Same passage chunks share the same color across different turn chunks.
\textit{(Best viewed in color)}} 
\label{figure:flow_example}
\end{figure}

Considering that the goal of the task is to generate interconnected questions in conversational question answering, CQG is challenging in a few aspects.
Firstly, a model should learn to generate conversational interconnected questions depending on the conversation so far. 
As shown in Table \ref{tab:example}, $\text{Q}_3$ is a single word `Why?', which should be `Why was he ineligible to serve a third term?' in a standalone interaction. Moreover, many questions in this conversation refer back to the conversation history using coreferences (e.g., $\text{Q}_2$, $\text{Q}_6$, $\text{Q}_9$), which is the nature of questions in a human conversation.
Secondly, a coherent conversation must have smooth transitions between turns (each turn contains a question-answer pair).
We expect the narrative structure of passages can influence the conversation flow of our interconnected questions.
We further investigate this point by conducting an analysis on our experiment dataset CoQA \cite{reddy2019coqa}.
We first split passages and turns of QA pairs into 10 uniform chunks and identify passage chunks of interest for each turn chunk.
Figure \ref{figure:flow_example} portrays the conversation flow between passage chunks and turn chunks. 
We see that in Figure \ref{figure:flow_example}, a question-answering style conversation usually starts focusing on the first few chunks in the passage and as the conversation advances, the focus shifts to the later passage chunks.

Previous works on question generation employ attentional sequence-to-sequence models on the crowd-sourced machine reading comprehension dataset SQuAD \cite{Rajpurkar2016SQuAD10}. 
They mainly focus on generating questions based on a single sentence (or paragraph) and an answer phrase \cite{Du2017LearningTA, Sun2018AnswerfocusedAP, Zhao2018ParagraphlevelNQ}, while in our setting, our model needs to not only ask a question on the given passage (paragraph) but also make the questions conversational by considering the conversation history.
Meanwhile, some researchers study question generation in dialogue systems to either achieve the correct answer through interactions \cite{Li2017LearningTD} or enhance the interactiveness and persistence of conversations \cite{Wang2018LearningTA}. Although questions in our setting are conversational, our work is different from these because our conversations are grounded in the given passages rather than open-domain dialogues.

\begin{figure*}[t!]
\centering
\includegraphics[width=1.0\textwidth]{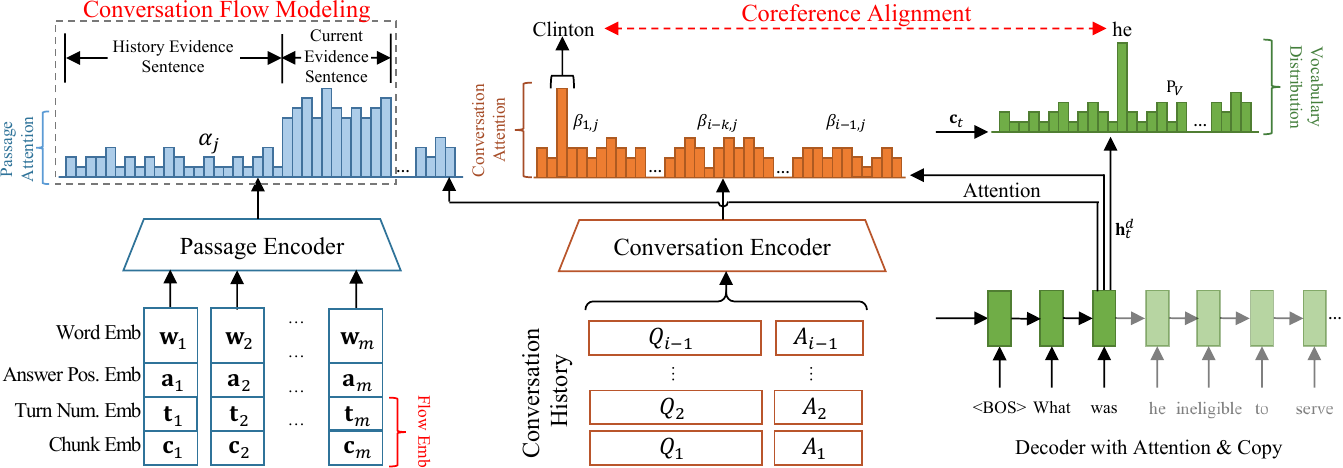}
\caption{The framework of our proposed model. For clarity, we omit to plot the copy mechanism in the figure. \textit{(Best viewed in color)}}
\label{figure:framework}
\end{figure*}

We propose a framework based on the attentional encoder-decoder model \cite{Luong2015EffectiveAT} to address this task.
To generate conversational questions (first challenge), we propose a multi-source encoder to jointly encode the passage and the conversation so far. 
At each decoding timestep, our model can learn to focus more on the passage to generate content words or on the conversation history to make the question succinct.
Furthermore, our coreference alignment modeling explicitly aligns coreferent mentions in conversation history (e.g. \textit{Clinton} in $\text{Q}_1$ Table \ref{tab:example}) with corresponding pronominal references in generated questions (e.g. \textit{he} in $\text{Q}_2$), which makes generated questions interconnected to conversation history.
The coreference alignment is implemented by adding extra supervision to bias the attention probabilities through a loss function.
The loss function explicitly guides our model to resolve to the correct non-pronominal coreferent mentions in the attention distribution and generate the correct pronominal references in target questions.
To make the conversations coherent (second challenge), we propose to model the conversation flow to transit focus inside the passage smoothly across turns.
The conversation flow modeling achieves this goal via a flow embedding and a flow loss. 
The flow embedding conveys the correlations between number of turns and narrative structure of passages. 
The flow loss explicitly encourages our model to focus on sentences contain key information to generate the current turn question and ignore sentences questioned several turns ago.

In evaluations on a conversational question answering dataset CoQA \cite{reddy2019coqa}, we find that our proposed framework outperforms several baselines in both automatic and human evaluations.
Moreover, the coreference alignment can greatly improve the precision and recall of generated pronominal references.
The conversation flow modeling can learn the smooth transition of conversation flow across turns.

\section{Problem Setting}
In this section, we define the Conversation Question Generation (CQG) task.
Given a passage $P$, a conversation history $C_{i-1} = \{\text{(}Q_{1}, A_{1}\text{)}, ..., \text{(}Q_{i-1},$  $A_{i-1}\text{)}\}$ and the aspect to ask (the current answer $A_{i}$), the task of CQG is to generate a question $\overline{Q_i}$ for the next turn:
\begin{align}
    \overline{Q}_i = \argmax_{Q_i} \text{Prob(}Q_i|P, A_i, C_{i-1}\text{)},
\end{align}
in which the generated question should be as conversational as possible.

Note that we formalize this setting as an answer-aware QG problem \cite{Zhao2018ParagraphlevelNQ}, which assumes answer phrases are given before generating questions. Moreover, answer phrases are shown as text fragments in passages. Similar problems have been addressed in \cite{Du2018HarvestingPQ, Zhao2018ParagraphlevelNQ, Sun2018AnswerfocusedAP}. 
Our problem setting can also be generalized to the answer-ignorant case. Models can identify which answers to ask first by combining question-worthy phrases extraction methods \cite{Du2018HarvestingPQ, Wang2019AMC}.

\section{Model Description}
As shown in Figure \ref{figure:framework}, our framework consists of four components: (1) multi-source encoder; (2) decoder with copy mechanism; (3) coreference alignment; (4) conversation flow modeling.

\subsection{Multi-Source Encoder} \label{sec.enc} 
Since a conversational question is dependent on a certain aspect of the passage $P$ and the conversation context $C_{i-1}$ so far, we jointly encode information from two sources via a passage encoder and a conversation encoder.

\paragraph{Passage Encoder.}
The passage encoder is a bidirectional-LSTM (bi-LSTM) \cite{Hochreiter1997LongSM}, which takes the concatenation of word embeddings $\mathbf{w}$ and answer position embeddings $\mathbf{a}$ as input $\mathbf{x}_i = [\mathbf{w}_i;\mathbf{a}_i]$.
We denote the answer span using the typical BIO tagging scheme and map each token in the paragraph into the corresponding answer position embedding (i.e., \texttt{B\_ANS}, \texttt{I\_ANS}, \texttt{O}).
Then the whole passage can be represented using the hidden states of the bi-LSTM encoder, i.e.,  $(\mathbf{h}^{p}_1, ..., \mathbf{h}^{p}_m)$, where $m$ is the sequence length.

\paragraph{Conversation Encoder.}
The conversation history $C_{i-1}$ is a sequence of question-answer pairs $\{\text{(}Q_{1}, A_{1}\text{)}, ...,\text{(}Q_{i-1}, A_{i-1}\text{)}\}$.
We use segmenters \textless$q$\textgreater \textless$a$\textgreater to concatenate each question answer pair $\text{(}Q, A\text{)}$ into a sequence of tokens (\textless$q$\textgreater, $q_{1}, ..., q_{m};$ \textless$a$\textgreater, $a_{1}, ..., a_{m}$).
We design a hierarchical structure to conduct conversation history modeling.
We first employ a token level bi-LSTM to get contextualized representation of question-answer pairs $(\mathbf{h}^{w}_{i-k, 1}, ..., \mathbf{h}^{w}_{i-k, m})$, where $i-k$ is the turn number and $k \in [1,i)$.
To model the dependencies across turns in the conversation history, we adopt a context level bi-LSTM to learn the contextual dependency $(\mathbf{h}^{c}_{1}, ..., \mathbf{h}^{c}_{i-1})$ across different turns (denoted in the subscript $1, ..., i-1$) of question-answer pairs. 

\subsection{Decoder with Attention \& Copy} \label{sec.dec}
The decoder is another LSTM to predict the word probability distribution.
At each decoding timestep $t$, it reads the word embedding $\mathbf{w}_{t}$ and the hidden state of previous timestep $\mathbf{h}^d_{t-1}$ to generate the current hidden state $\mathbf{h}^d_t = \text{LSTM(}\mathbf{w}_t, \mathbf{h}^d_{t-1}\text{)}$.

To generate a conversational question grounded in the passage, the decoder itself should decide to focus more on passage hidden states $\mathbf{h}^p_j$ or the hidden states of conversation history $\mathbf{h}^{w}_{i-k,j}$ at each decoding timestep.
Therefore, we flat token level conversation hidden states $\mathbf{h}^{w}_{i, j}$ and aggregate the passage hidden states $\mathbf{h}^{p}_{j}$ with token level conversation hidden states $\mathbf{h}^{w}_{i, j}$ into a unified memory: $\text{(}\mathbf{h}^{p}_{1}, ..., \mathbf{h}^{p}_{m}$; $\mathbf{h}^{w}_{1, 1}, ..., \mathbf{h}^{w}_{1, m}$; ... ; $\mathbf{h}^{w}_{i-1, 1}, ..., \mathbf{h}^{w}_{i-1, m}\text{)}$, where $\mathbf{h}^{w}_{i, j}$ denotes the \textit{j}-th token of the \textit{i}-th turn in token level conversation hidden states. 
Then we attend the unified memory with the standard attention mechanism \cite{Luong2015EffectiveAT} for the passage attention $\text{(}\alpha_1, ..., \alpha_m\text{)}$ and the hierarchical attention mechanism for the conversation attention $\text{(}\beta_{1, 1}, ..., \beta_{1, m}$; ...; $\beta_{i-1, 1}, ..., \beta_{i-1, m}\text{)}$:
\begin{align}
e^p_{j} &= {\mathbf{h}^p_j}^\top \mathbf{W}_p \mathbf{h}^d_{t}, \\
e^{w}_{i-k,j} &= {\mathbf{h}^{w}_{i-k,j}}^\top \mathbf{W}_{w} \mathbf{h}^d_{t}, \\
e^{c}_{i-k} &= {\mathbf{h}^{c}_{i-k}}^\top \mathbf{W}_{c} \mathbf{h}^d_{t}, \\
\alpha_j = \frac{e^p_{j}}{e_\text{total}},& {~~}
\beta_{i-k, j} = \frac{e^{w}_{i-k,j} * e^{c}_{i-k}}{e_\text{total}}, \label{eqn.hieratt} 
\end{align}
where $e_{\text{total}} = \Sigma_{j}e^p_{j} + \Sigma_{k,j}{e^{w}_{i-k,j} * e^{c}_{i-k}}$ and $\mathbf{W}_p$, $\mathbf{W}_{w}$, $\mathbf{W}_{c}$ are learnable weights.

Finally, we derive the context vector $\mathbf{c}_t$ and the final vocabulary distribution $\text{P}_{V}$:
\begin{align}
\mathbf{c}_t &= \Sigma_{j}\alpha_{j} \mathbf{h}^p_j + \Sigma_{j,k} \beta_{i-k, j} \mathbf{h}^{w}_{i-k,j}\nonumber, \\
\text{P}_{V} &= \text{softmax(} \mathbf{W}_v \text{(tanh(} \mathbf{W}_a [\mathbf{h}^d_{t};\mathbf{c}_t] \text{)} + \mathbf{b}_v \text{)} \nonumber,
\end{align}
where $\mathbf{W}_v$, $\mathbf{W}_a$ are learnable weights.
Please refer to \newcite{See2017GetTT} for more details on the copy mechanism.

\subsection{Coreference Alignment} \label{sec.coref}
Using coreferences to refer back is an essential property of conversational questions.
Almost half of the questions contains explicit coreference markers such as \textit{he}, \textit{she}, \textit{it} in CoQA \cite{reddy2019coqa}.
Therefore, we propose the coreference alignment to enable our model such ability.
Take $\text{Q}_2$ in Table \ref{tab:example} as an example, traditional question generation system can only generate question like ``What was \textit{Clinton} ineligible to serve?'', while our system with coreference alignment can align the name ``\textit{Clinton}'' to its pronominal reference ``\textit{he}'' and generate a more conversational question ``What was \textit{he} ineligible to serve?''.

The coreference alignment modeling tells the decoder to look at the correct non-pronominal coreferent mention in the conversation attention distribution to produce the pronominal reference word.
We achieve this via two stages.
In the pre-processing stage, given the conversation history $C_{i-1}$ and the question $Q_i$ which has a pronominal reference (e.g., \textit{he} for $\text{Q}_2$ in Table \ref{tab:example}), we first run a coreference resolution system \cite{Clark2016DeepRL} to find its coreferent mention $(w^c_1, ... w^c_m)$ (e.g. \textit{Clinton}) in the conversation history $C_{i-1}$, where the superscript $c$ denotes tokens identified as the coreferent mention.
During training, we introduce a novel loss function built on the conversation attention of coreferent mentions $\beta^c_i$ and the output word probability of its pronominal reference word $p_\text{coref} \in \text{P}_V$. 
As shown in Figure \ref{figure:framework}, when our model need to refer back to the coreferent mention, we ask the model focus correctly on the antecedent (e.g. \textit{Clinton}) and maximize the probability of its pronominal reference (e.g. \textit{he}) $p_\text{coref}$ in the output vocabulary distribution $\text{P}_V$,
\begin{gather}
\mathcal{L}_\text{coref} = - ( \lambda_1 {\log}\frac{\Sigma_{j}{\beta^c_j}}{\Sigma_{k,j}{\beta_{i-k,j}}} + \lambda_2 {\log}{p_\text{coref}} ) * s_c\nonumber,
\end{gather}
where $\lambda_1, \lambda_2$ are hyperparameters, $s_c$ is the confidence score between the non-pronominal coreferent mention and the pronoun obtained during the pre-processing stage.

\subsection{Conversation Flow Modeling} \label{sec.flow}
Another key challenge in CQG is that a coherent conversation must have smooth transitions between turns. As illustrated in Figure \ref{figure:flow_example}, we find that as the conversations go on, most of the questioners transit their focus from the beginning of passages to the end.
Following this direction, we model the conversation flow to learn smooth transitions across turns of the conversation.

\paragraph{Flow Embedding.} 
As shown in Figure \ref{figure:framework}, we feed our model with the current turn number indicator in the conversation and the relative position for each token in the passage, which, intuitively, are useful for modeling the conversation flow.
We achieve this goal via two additional embeddings. 
The turn number embedding is a learned lookup table $[\mathbf{t}_1, ..., \mathbf{t}_n]$ to map the turn number $i$ into its feature embedding space, where $n$ is the maximum turn we consider. 
For encoding the relative position of each token, we split the passage into $L$ uniform chunks. Each token in the passage is mapped to its corresponding chunk embedding $[\mathbf{c}_1, ..., \mathbf{c}_L]$. 
The final input to the passage encoder is the concatenation of word embedding, answer position embedding (introduced in Section \ref{sec.enc}) and these two additional embeddings: $\mathbf{x}_i = [\mathbf{w}_i; \mathbf{a}_i; \mathbf{t}_i; \mathbf{c}_i]$.

We further add a gated self-attention modeling mechanism \cite{Zhao2018ParagraphlevelNQ} in the passage encoder.
Motivating our use of self-attention we consider two desiderata.
One is self-attention with answer position embedding can aggregate answer-relevant information from the whole passage for question generation. Another is we want to learn the latent alignment between the turn number embedding and the chunk embedding for better modeling the conversation flow.
We first match the rich-feature enhanced passage representation $\mathbf{H}^p = [\mathbf{h}^p_1; ...; \mathbf{h}^p_m]$ with itself $\mathbf{h}^p_j$ to compute the self-matching representation $\mathbf{u}^p_j$, and then combine it with the original representation $\mathbf{h}^p_j$:
\begin{align}
    \mathbf{a}^p_j = \text{softm} & \text{ax(}{\mathbf{H}^p}^\top \mathbf{W}_{s} \mathbf{h}^p_j \text{)}, 
    ~\mathbf{u}^p_j = \mathbf{H}^p \mathbf{a}^p_j \\
    \mathbf{f}^p_j &= \text{tanh(}\mathbf{W}_f[\mathbf{h}^p_j; \mathbf{u}^p_j]\text{)},
\end{align}
The final representation $\tilde{\mathbf{h}}^p_j$ is derived via a gated summation through a learnable gate vector $\mathbf{g}^p_j$,
\begin{align}
    \mathbf{g}^p_t &= \text{sigmoid(}\mathbf{W}_g[\mathbf{h}^p_j; \mathbf{u}^p_j]\text{)} \\
    \tilde{\mathbf{h}}^p_j &= \mathbf{g}^p_t \odot \mathbf{f}^p_j + (1-\mathbf{g}^p_t) \odot \mathbf{h}^p_j
\end{align}
where $\mathbf{W}_{s}$, $\mathbf{W}_{f}$, $\mathbf{W}_{g}$ are learnable weights, $\odot$ is the element-wise multiplication. Self matching enhanced representation $\tilde{\mathbf{h}}^p_j$ takes the place of the passage representation $\mathbf{h}^p_j$ for calculating the passage attention.

\paragraph{Flow Loss.}
In Section \ref{sec.enc}, our answer position embedding can help model the conversation flow by showing the position of answer fragments inside the passage.
However, it is still helpful to tell the model explicitly which sentences around the answer are of high informativity to generate the current turn question. 
The flow loss is designed to help our model to locate the evidence sentences correctly. 
Firstly, we define two kinds of sentences in the passage. 
If a sentence is informative to the current question, we call it \texttt{Current Evidence Sentence (CES)}. 
If a sentence is informative to questions in the conversation history and irrelevant to the current question, we call it \texttt{History Evidence Sentence (HES)}.
Then our model is taught to focus on current evidence sentences and ignore the history evidence sentences in the passage attention $\alpha_j$ via the following flow loss:
\begin{gather}\label{eqn.flow}
    \mathcal{L}_\text{flow} 
    = - \lambda_3 \text{log} \frac{\Sigma_{j: w_j \in \texttt{CES}}\alpha_j}{\Sigma_j \alpha_j} + \lambda_4 \frac{\Sigma_{j: w_j \in \texttt{HES}}\alpha_j}{\Sigma_j \alpha_j} \nonumber
\end{gather}
where $\lambda_3, \lambda_4$ are hyperparameters, and $w_j \in \texttt{CES}/\texttt{HES}$ indicates the token $w_j$ is inside the sentence with a \texttt{CES}/\texttt{HES} label.

\subsection{Joint Training} \label{sec.full}
Considering all the aforementioned components, we define a joint loss function as:
\begin{align}
    \mathcal{L} = \mathcal{L}_\text{nll} + \mathcal{L}_\text{coref} + \mathcal{L}_\text{flow},
\end{align}
in which $\mathcal{L}_\text{nll} = - \text{log Prob(}Q_i|P, A_i, C_{i-1}\text{)}$ is the the negative log-likelihood loss in the sequence to sequence learning \cite{Sutskever2014SequenceTS}.

\section{Experiments}

\subsection{Dataset Preparation}\label{sec.dataset}
We conduct experiments on the CoQA dataset \cite{reddy2019coqa}.
It is a large-scale conversational question answering dataset for measuring the ability of machines to participate in a question-answering style conversation.
The authors employ Amazon Mechanical Turk to collect 8k conversations with 127k QA pairs. 
Specifically, they pair two crowd-workers: a questioner and an answerer to chat about a passage. 
The answerers are asked to firstly highlight extractive spans in the passage as rationales and then write the free-form answers.
We first extract each data sample as a quadruple of passage, question, answer and conversation history (previous $n$ turns of QA pairs) from CoQA.
Then we filter out QA pairs with \textit{yes}, \textit{no} or \textit{unknown} as answers (28.7\% of total QA pairs) because there is too little information to generate the question to the point. 
Finally, we randomly split the dataset into a training set (80\%, 66298 samples), a validation set (10\%, 8409 samples) and a testing set  (10\%, 8360 samples). The average passage, question and answer lengths are 332.9, 6.3 and 3.2 tokens respectively. 

\subsection{Implementation Details}
\paragraph{Locating Extractive Answer Spans.}
As studied by \newcite{Yatskar2018AQC}, abstractive answers in CoQA are mostly small modifications to spans occurring in the context. 
The maximum achievable performance by a model that predicts spans from the context is 97.8 F1 score.
Therefore, we find the extractive spans from the passage which have the maximum F1 score with answers and treat them as answers for our answer position embedding.

\paragraph{Number of Turns in Conversation History.}
\newcite{reddy2019coqa} find that in CoQA dataset, most questions in a conversation have a limited dependency within a bound of two turns. Therefore, we choose the number of history turns as $n=3$ to ensure the target questions have enough conversation history information to generate and avoid introducing too much noise from all turns of QA pairs.

\paragraph{Labeling Evidence Sentences.}
As mentioned in Section \ref{sec.dataset}, the crowd-workers label the extractive spans in the passage as rationales for actual answers. 
We treat sentences containing the rationale as \texttt{Current Evidence Sentence}.

\paragraph{Model Settings.}
We employ the teacher-forcing training, and in the generating stage, we set the maximum length for output sequence as 15 and block unigram repeated token, the beam size $k$ is set to 5. 
All hyperparameters and models are selected on the validation set and the results are reported on the test set.

\subsection{Baselines and Ablations}
We compare with the state-of-the-art baselines and conduct ablations as follows: \textbf{PGNet} is the pointer-generator network \cite{See2017GetTT}. We concatenate the passage $P$, the conversation history $C_{i-1}$ and the current answer $A_i$ as a sequence for the input.
\textbf{NQG} \cite{Du2018HarvestingPQ} is similar to the previous one but it takes current answer features concatenated with the word embeddings during encoding. 
\textbf{MSNet} is our base model \textbf{M}ulti-\textbf{S}ource encoder decoder network (Section \ref{sec.enc} \& \ref{sec.dec}).
\textbf{CorefNet} is our proposed \textbf{Coref}erence alignment model (Section \ref{sec.coref}).
\textbf{FlowNet} is our proposed conversation \textbf{Flow} model (Section \ref{sec.flow}).
\textbf{CFNet} is the model with both the \textbf{C}oreference alignment and the conversation \textbf{F}low modeling.

\section{Results and Analysis}
\subsection{Main Results}\label{sec.result.main}
Since the average length of questions is 6.3 tokens only, we employ BLEU (1-3) \cite{Papineni2002BleuAM} and ROUGE-L (R-L) \cite{Lin2004ROUGEAP} scores to evaluate n-gram similarity between the generated questions with the ground truth.
We evaluate baselines and our models by predicting the current question given a passage, the current answer, and the ground truth conversation history. 
\begin{table}[!t]
    \small
    \centering
    \resizebox{0.8\columnwidth}{!}{
    \begin{tabular}{@{}l@{~} @{~}l@{~} | @{~}c@{~} @{~}c@{~} @{~}c@{~} @{~}c@{~}}
    \Xhline{3\arrayrulewidth}
   & & B1 & B2 & B3 & R-L  \\
    \hline\hline
   \multirow{2}{*}{}  
    &  PGNet & 28.84* & 13.74* & {~}{~}8.16* & 39.18* \\
    &  NQG   & 35.56* & 21.14* & 14.84* & 45.58*   \\
    \hline
    &  MSNet & 36.27* & 21.92* & 15.51* & 46.01*   \\
    &  CorefNet  & \underline{36.89}{~}{~} & \underline{22.28}{~}{~} & \underline{15.77}{~}{~} & \underline{46.53}{~}{~} \\
    &  FlowNet   & \underline{36.87}{~}{~} & 22.49{~}{~} & 15.98{~}{~} & 46.64{~}{~}  \\
    &  CFNet & \textbf{37.38}{~}{~} & \textbf{22.81}{~}{~} & \textbf{16.25}{~}{~} &  \textbf{46.90}{~}{~}  \\
    \Xhline{3\arrayrulewidth}
    \end{tabular}}
    \caption{Main results of baselines and our models. \textit{t}-test is conducted between our CFNet and baselines/ablations. (\underline{underline}: \textit{p-value} \textless 0.05, *: \textit{p-value} \textless 0.01).}
    \label{tab:result-main}
\end{table}

Table \ref{tab:result-main} shows the main results, and we have the following observations:
\begin{itemize}[leftmargin=*]
\setlength{\itemsep}{0pt}
\setlength{\parsep}{0pt}
\setlength{\parskip}{0pt}
\item NQG outperforms PGNet by a large margin. 
The improvement shows that the answer position embedding \cite{Zhou2017NeuralQG} is helpful for asking questions to the point.
\item Our base model MSNet outperforms NQG, which reveals that the hierarchical encoding and the hierarchical attention to conversation history can model the dependency across different turns in conversations.
\item Both our CorefNet and FlowNet outperform our base model. We will analyze the effectiveness of our coreference alignment and conversation flow modeling in the following two sections respectively.
\item Our CFNet is significantly better than two baselines (PGNet, NQG), our MSNet, and our CorefNet. However, the difference between our CFNet and our FlowNet is not significant. This is because the conversation flow modeling improves all test samples while the coreference alignment contributes only to questions containing pronominal references.
\end{itemize}

\subsection{Coreference Alignment Analysis}\label{sec.result.coref}
As we discussed in Section \ref{sec.coref}, it is the nature of conversational questions to use coreferences to refer back.
In order to demonstrate the effectiveness of the proposed coreference alignment, we evaluate models on a subset of the test set called coreference set. 
Each sample in the coreference set requires a pronoun resolution between the conversation history and the current question (e.g., $\text{Q}_2$, $\text{Q}_6$, $\text{Q}_9$ in Table \ref{tab:example}). 
In additional to the BLEU(1-3) and ROUGE-L metrics, we also calculate the Precision (P), Recall (R) and F-score (F) of pronouns in the generated questions with regard to pronouns in the ground truth questions.

\begin{table}[t]
    \centering
    \resizebox{1.0\columnwidth}{!}{
    \begin{tabular}{@{}l@{~} | @{~}c@{~} @{~}c@{~} @{~}c@{~} @{~}c@{~} | @{~}c@{~} @{~}c@{~} @{~}c@{~} } 
    \Xhline{3\arrayrulewidth}
    \hline
    & B1 & B2 & B3 & R-L & P & R & F\\
      \hline\hline
    PGNet & 27.66* & 13.82* & {~}{~}8.96*  & 38.40* & 26.87* & 25.17* & 25.68*  \\
    NQG   & 34.75* & 21.52* & 15.96* & 45.04* & 34.46* & 32.97* & 33.25*  \\
    MSNet & 36.31* & \underline{22.92}{~}{~} & \underline{17.07}{~}{~} & 45.97* & 35.34* & 33.80* & 34.07*  \\
    \cline{1-8}
    CorefNet & \textbf{37.51}{~}{~} & \textbf{24.14}{~}{~} & \textbf{18.44}{~}{~} & \textbf{47.45}{~}{~} & \textbf{42.09}{~}{~} & \textbf{40.35}{~}{~} & \textbf{40.64}{~}{~} \\
    \Xhline{3\arrayrulewidth}
    \end{tabular}}
    \caption{Evaluation results on the coreference test set. Precision (P), Recall (R) and F-score (F) of predicted pronouns are also reported. Significant tests with \textit{t}-test are conducted between CorefNet and models without the coreference alignment. (\underline{underline}: \textit{p-value} \textless 0.05, *: \textit{p-value} \textless 0.01).}
    \label{tab:result-coref}
\end{table}

\begin{figure}[t!]
\centering
\includegraphics[width=1.0\columnwidth]{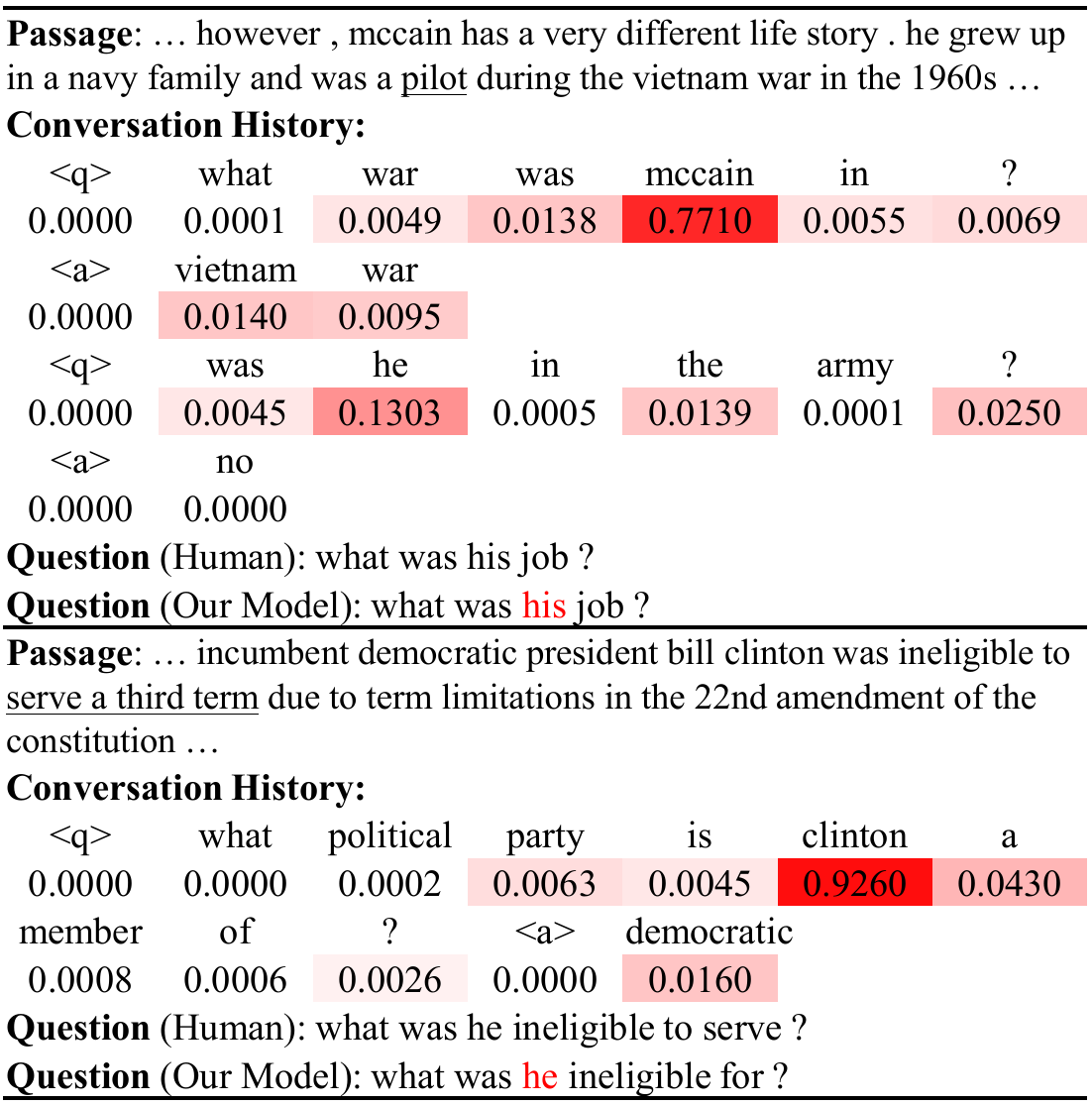}
\caption{Examples for the coreference alignment model. We show the attention probability (renormalize to 1) when the CorefNet predicts a pronoun (red color in Question). The current answers are underlined in the passages. \textit{(Best viewed in color)}}
\label{figure:coref_case}
\end{figure}

The results are depicted in Table \ref{tab:result-coref}. 
With the help of the coreference alignment, CorefNet significantly improves the precision, recall, and f-score of the predicted pronouns.
Moreover, the performance on n-gram overlapping metrics is also boosted.
To gain more insights into how the coreference alignment model influence the generation process, in Figure \ref{figure:coref_case}, we visualize the conversation attention distribution $\beta_j$ at the timestep the model predicts a pronoun.
The conversation history distribution $\beta_j$ is renormalized to $\Sigma_j \beta_j = 1$.
All two examples show that our model put the highest attention probability on the coreferent mentions (i.e. \textit{McCain/Clinton}) when it generates the pronominal references (\textit{his/he}).
We can conclude that our coreference alignment model can align correct coreferent mentions to generate corresponding pronouns.

\begin{figure*}[t!]
\centering
\includegraphics[width=1.0\textwidth]{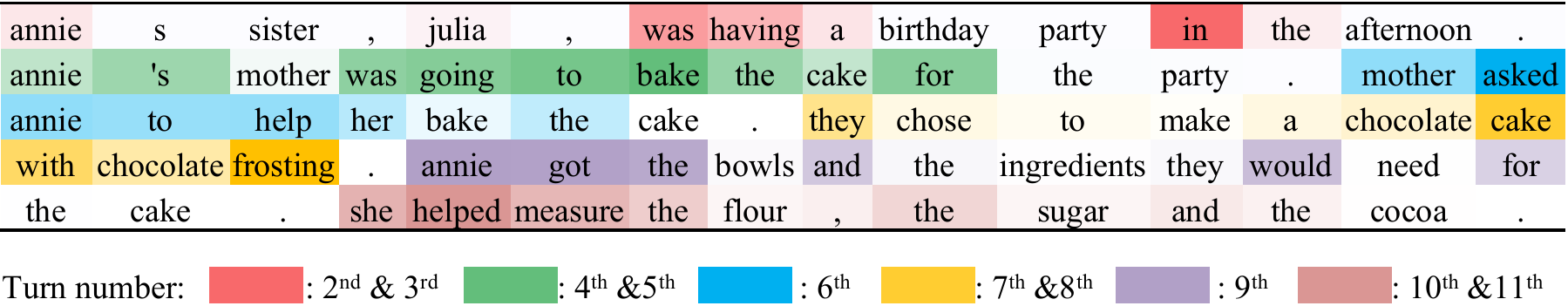}
\caption{The transition of passage attention distribution between turns. Different colors are correspond to different turns. To show attention probability of different turns in one place, we only draw attention probability $\alpha_j$ \textgreater 0.1 here. If two turns focus on the same sentence, we average the attention probability between them. \textit{(Best viewed in color)}}
\label{figure:flow_case}
\end{figure*}

\subsection{Conversation Flow Modeling Analysis} \label{sec.result.flow}
As discussed in Section \ref{sec.flow}, a coherent conversation should have smooth transitions between turns, and we design our model to follow the narrative structure of the passage. 
Figure \ref{figure:flow_case} shows an example illustrating the transition of passage attention distribution $a_j$ (normalize to 1) during first 11 turns of a conversation.
We see that the model transits its focus smoothly across the first 11 turns from the first sentence in the passage to later parts.
Sometimes the model drills down with two questions for the same sentence such as turn 2 \& 3, 4 \& 5 and 10 \& 11.

To quantitatively validate the effectiveness of our conversation flow modeling, we study the alignment between passage attention $\alpha_j$ and sentences of interest in the passage.
Ideally, a successful model should focus on sentences of interest (i.e., Current Evidence Sentence) and ignore sentences questioned several turns ago (i.e., History Evidence Sentence).
We validate this intuition by calculating $\Sigma_{j: w_j \in \texttt{CES}}\alpha_j$ and $\Sigma_{j: w_j \in \texttt{HES}}\alpha_j$ for all examples in test set.
Results show that $\Sigma_{j: w_j \in \texttt{CES}}\alpha_j$ and $\Sigma_{j: w_j \in \texttt{HES}}\alpha_j$ for our model with conversation flow modeling are 0.9966 and 0.0010 on average, which demonstrates that our conversation flow modeling can locate the current evidence sentences precisely and ignore the history evidence sentence.
For the model without the flow modeling (CorefNet), $\Sigma_{j: w_j \in \texttt{CES}}\alpha_j = 0.4093$, $\Sigma_{j: w_j \in \texttt{HES}}\alpha_j = 0.1778$, which proves our intuition in Section \ref{sec.flow} that the answer position embedding cannot have comparable effects on the conversation flow modeling.

\subsection{Human Evaluation}
We randomly sample 93 questions with the associated passage and conversation history to conduct human evaluation. We hire 5 workers to evaluate the questions generated by PGNet, MSNet, and our CFNet.
All models are evaluated in terms of following 3 metrics: ``Grammaticality'', ``Answerability'' and ``Interconnectedness''.
``Grammaticality'' measures the grammatical correctness and fluency of the generated questions. 
``Answerability'' evaluates whether the generated question can be answered by the current answer. 
``Interconnectedness'' measures whether the generated questions are \textit{conversational} or not. If a question refers back to the conversation history using coreference or is dependent on the conversation history such as incomplete questions `Why?', `Of what?', we define it as a \textit{conversational} question. All metrics are rated on a 1-3 scale (3 for the best).

\begin{table}[t]
    \centering
    \resizebox{0.98\columnwidth}{!}{
    \begin{tabular}{@{}l@{~} | @{~}c@{~} @{~}c@{~} @{~}c@{~} } 
    \Xhline{3\arrayrulewidth}
    \hline
    & Grammaticality & Answerability & Interconnectedness \\
    \hline
    PGNet  & 2.74 & 1.39{~}{~} & 1.59{~}{~}  \\
    MSNet  & 2.85 & 2.39{~}{~} & 1.74{~}{~} \\
    CFNet  & 2.89 & \textbf{2.74}* & \textbf{2.67}*  \\
    \Xhline{3\arrayrulewidth}
    \end{tabular}}
    \caption{Manual evaluation results. All metrics are rated on a 1-3 scale (3 for the best). Two-tailed \textit{t}-test results are shown for our CFNet compared to PGNet/MSNet. * indicates \textit{p-value} \textless 0.01.}
    \label{tab:result-human}
\end{table}

The results are shown in Table \ref{tab:result-human}. 
All models achieve high scores on ``Grammaticality'', owing to the strong language modeling capability of neural models.
MSNet and our CFNet perform well on ``Answerability'' while PGNet does not. This demonstrates our base model MSNet and our CFNet can ask questions to the point.
Finally, our CFNet outperforms the other two models in terms of ``Interconnectedness'' by a large gap, which proves that the proposed coreference alignment and conversation flow modeling can effectively make questions \textit{conversational}.

\section{Related Work}
The task of Question Generation (QG) aims at generating natural questions from given input contexts.
Some template-based approaches \cite{Vanderwende2007AnsweringAQ,Heilman2010GoodQS} were proposed initially, where well-designed rules and heavy human labor are required for declarative-to-interrogative sentence transformation. 
With the rise of data-driven learning approach and sequence to sequence (seq2seq) framework \cite{Sutskever2014SequenceTS}, \newcite{Du2017LearningTA} first formulate QG as a seq2seq problem with attention mechanism.
They extract sentences and pair them with questions from SQuAD \cite{Rajpurkar2016SQuAD10}, a large-scale reading comprehension dataset.
Recent works along this line focus on how to utilize the answer information better to generate questions to the point \cite{Zhou2017NeuralQG,Gao2019GeneratingDF,Sun2018AnswerfocusedAP}, how to generate questions with specific difficulty levels \cite{Gao2018DifficultyCQ} and how to effectively use the contexts in paragraphs to generate questions that cover context beyond a single sentence \cite{Zhao2018ParagraphlevelNQ, Du2018HarvestingPQ}.

In parallel to question generation for reading comprehension, some researchers recently investigate question generation in dialogue systems.
\newcite{Li2017LearningTD} show that asking questions through interactions can receive useful feedbacks to reach the correct answer. 
\newcite{Wang2018LearningTA} consider asking questions in open-domain conversational systems with typed decoders to enhance the interactiveness and persistence of conversations. 

In this paper, we propose a new setting which is related to the above two lines of research. 
We consider asking questions grounded in a passage via a question-answering style conversation. Since the questions and answers are in the format of a conversation, questions in our setting are highly conversational and interconnected to conversation history.
This setting is challenging because we need to jointly model the attention shifting in the passage and the structure of a conversation \cite{Grosz1986AttentionIA}.
A limitation of the conversation in our setting is that we can only generate a series of interconnected questions according to predefined answers but in a real dialog the questioner can ask different questions according to the answers' response.

\section{Conclusion and Future Work}
In this paper, we study the problem of question-answering style Conversational Question Generation (CQG), which has never been investigated before. 
We propose an end-to-end neural model with coreference alignment and conversation flow modeling to solve this problem.
The coreference alignment enables our framework to refer back to the conversation history using coreferences.
The conversation flow modeling builds a coherent conversation between turns.
Experiments show that our proposed framework achieves the best performance in automatic and human evaluations.

There are several future directions for this setting. 
First, the presented system is still contingent on highlighting answer-like nuggets in the declarative text. Integrating answer span identification into the presented system is a promising direction.
Second, in our setting, the roles of the questioner and the answerer are fixed. However, questions can be raised by either part in real scenario.

\section*{Acknowledgments}
This work is supported by the Research Grants Council of the Hong Kong
Special Administrative Region, China (No. CUHK 14208815 and No. CUHK
14210717 of the General Research Fund).
We thank Department of Computer Science and Engineering, The Chinese University of Hong Kong for the conference grant support.
We would like to thank Wang Chen and Jingjing Li for their comments.

\bibliography{acl2019}

\begin{thebibliography}{27}
\expandafter\ifx\csname natexlab\endcsname\relax\def\natexlab#1{#1}\fi

\bibitem[{Clark and Manning(2016)}]{Clark2016DeepRL}
Kevin Clark and Christopher~D. Manning. 2016.
\newblock \href {https://doi.org/10.18653/v1/D16-1245} {Deep reinforcement
  learning for mention-ranking coreference models}.
\newblock In \emph{Proceedings of the 2016 Conference on Empirical Methods in
  Natural Language Processing}, pages 2256--2262, Austin, Texas. Association
  for Computational Linguistics.

\bibitem[{Du and Cardie(2018)}]{Du2018HarvestingPQ}
Xinya Du and Claire Cardie. 2018.
\newblock \href {https://www.aclweb.org/anthology/P18-1177} {Harvesting
  paragraph-level question-answer pairs from {W}ikipedia}.
\newblock In \emph{Proceedings of the 56th Annual Meeting of the Association
  for Computational Linguistics (Volume 1: Long Papers)}, pages 1907--1917,
  Melbourne, Australia. Association for Computational Linguistics.

\bibitem[{Du et~al.(2017)Du, Shao, and Cardie}]{Du2017LearningTA}
Xinya Du, Junru Shao, and Claire Cardie. 2017.
\newblock \href {https://doi.org/10.18653/v1/P17-1123} {Learning to ask: Neural
  question generation for reading comprehension}.
\newblock In \emph{Proceedings of the 55th Annual Meeting of the Association
  for Computational Linguistics (Volume 1: Long Papers)}, pages 1342--1352,
  Vancouver, Canada. Association for Computational Linguistics.

\bibitem[{Gao et~al.(2019{\natexlab{a}})Gao, Bing, Chen, Lyu, and
  King}]{Gao2018DifficultyCQ}
Yifan Gao, Lidong Bing, Wang Chen, Michael~R. Lyu, and Irwin King.
  2019{\natexlab{a}}.
\newblock Difficulty controllable generation of reading comprehension
  questions.
\newblock In \emph{Proceedings of the Twenty-Eightth International Joint
  Conference on Artificial Intelligence, {IJCAI-19}}. International Joint
  Conferences on Artificial Intelligence Organization.

\bibitem[{Gao et~al.(2019{\natexlab{b}})Gao, Bing, Li, King, and
  Lyu}]{Gao2019GeneratingDF}
Yifan Gao, Lidong Bing, Piji Li, Irwin King, and Michael~R. Lyu.
  2019{\natexlab{b}}.
\newblock Generating distractors for reading comprehension questions from real
  examinations.
\newblock In \emph{AAAI Conference on Artificial Intelligence}.

\bibitem[{Grosz and Sidner(1986)}]{Grosz1986AttentionIA}
Barbara~J. Grosz and Candace~L. Sidner. 1986.
\newblock \href {https://www.aclweb.org/anthology/J86-3001} {Attention,
  intentions, and the structure of discourse}.
\newblock \emph{Computational Linguistics}, 12(3):175--204.

\bibitem[{Heilman and Smith(2010)}]{Heilman2010GoodQS}
Michael Heilman and Noah~A. Smith. 2010.
\newblock \href {https://www.aclweb.org/anthology/N10-1086} {Good question!
  statistical ranking for question generation}.
\newblock In \emph{Human Language Technologies: The 2010 Annual Conference of
  the North {A}merican Chapter of the Association for Computational
  Linguistics}, pages 609--617, Los Angeles, California. Association for
  Computational Linguistics.

\bibitem[{Hochreiter and Schmidhuber(1997)}]{Hochreiter1997LongSM}
Sepp Hochreiter and J{\"u}rgen Schmidhuber. 1997.
\newblock Long short-term memory.
\newblock \emph{Neural Computation}, 9:1735--1780.

\bibitem[{Li et~al.(2017)Li, Miller, Chopra, Ranzato, and
  Weston}]{Li2017LearningTD}
Jiwei Li, Alexander~H. Miller, Sumit Chopra, Marc'Aurelio Ranzato, and Jason
  Weston. 2017.
\newblock Learning through dialogue interactions by asking questions.
\newblock In \emph{ICLR}.

\bibitem[{Lin(2004)}]{Lin2004ROUGEAP}
Chin-Yew Lin. 2004.
\newblock \href {https://www.aclweb.org/anthology/W04-1013} {{ROUGE}: A package
  for automatic evaluation of summaries}.
\newblock In \emph{Text Summarization Branches Out: Proceedings of the {ACL}-04
  Workshop}, pages 74--81, Barcelona, Spain. Association for Computational
  Linguistics.

\bibitem[{Luong et~al.(2015)Luong, Pham, and Manning}]{Luong2015EffectiveAT}
Thang Luong, Hieu Pham, and Christopher~D. Manning. 2015.
\newblock \href {https://doi.org/10.18653/v1/D15-1166} {Effective approaches to
  attention-based neural machine translation}.
\newblock In \emph{Proceedings of the 2015 Conference on Empirical Methods in
  Natural Language Processing}, pages 1412--1421, Lisbon, Portugal. Association
  for Computational Linguistics.

\bibitem[{Mostafazadeh et~al.(2016)Mostafazadeh, Misra, Devlin, Mitchell, He,
  and Vanderwende}]{Mostafazadeh2016GeneratingNQ}
Nasrin Mostafazadeh, Ishan Misra, Jacob Devlin, Margaret Mitchell, Xiaodong He,
  and Lucy Vanderwende. 2016.
\newblock \href {https://doi.org/10.18653/v1/P16-1170} {Generating natural
  questions about an image}.
\newblock In \emph{Proceedings of the 54th Annual Meeting of the Association
  for Computational Linguistics (Volume 1: Long Papers)}, pages 1802--1813,
  Berlin, Germany. Association for Computational Linguistics.

\bibitem[{Papineni et~al.(2002)Papineni, Roukos, Ward, and
  Zhu}]{Papineni2002BleuAM}
Kishore Papineni, Salim Roukos, Todd Ward, and Wei-Jing Zhu. 2002.
\newblock \href {https://doi.org/10.3115/1073083.1073135} {{B}leu: a method for
  automatic evaluation of machine translation}.
\newblock In \emph{Proceedings of 40th Annual Meeting of the Association for
  Computational Linguistics}, pages 311--318, Philadelphia, Pennsylvania, USA.
  Association for Computational Linguistics.

\bibitem[{Rajpurkar et~al.(2016)Rajpurkar, Zhang, Lopyrev, and
  Liang}]{Rajpurkar2016SQuAD10}
Pranav Rajpurkar, Jian Zhang, Konstantin Lopyrev, and Percy Liang. 2016.
\newblock \href {https://doi.org/10.18653/v1/D16-1264} {{SQ}u{AD}: 100,000+
  questions for machine comprehension of text}.
\newblock In \emph{Proceedings of the 2016 Conference on Empirical Methods in
  Natural Language Processing}, pages 2383--2392, Austin, Texas. Association
  for Computational Linguistics.

\bibitem[{Reddy et~al.(2019)Reddy, Chen, and Manning}]{reddy2019coqa}
Siva Reddy, Danqi Chen, and Christopher~D Manning. 2019.
\newblock {CoQA}: A conversational question answering challenge.
\newblock \emph{Transactions of the Association for Computational Linguistics}.

\bibitem[{See et~al.(2017)See, Liu, and Manning}]{See2017GetTT}
Abigail See, Peter~J. Liu, and Christopher~D. Manning. 2017.
\newblock \href {https://doi.org/10.18653/v1/P17-1099} {Get to the point:
  Summarization with pointer-generator networks}.
\newblock In \emph{Proceedings of the 55th Annual Meeting of the Association
  for Computational Linguistics (Volume 1: Long Papers)}, pages 1073--1083,
  Vancouver, Canada. Association for Computational Linguistics.

\bibitem[{Serban et~al.(2016)Serban, Garc{\'\i}a-Dur{\'a}n, Gulcehre, Ahn,
  Chandar, Courville, and Bengio}]{Serban2016GeneratingFQ}
Iulian~Vlad Serban, Alberto Garc{\'\i}a-Dur{\'a}n, Caglar Gulcehre, Sungjin
  Ahn, Sarath Chandar, Aaron Courville, and Yoshua Bengio. 2016.
\newblock \href {https://doi.org/10.18653/v1/P16-1056} {Generating factoid
  questions with recurrent neural networks: The 30{M} factoid question-answer
  corpus}.
\newblock In \emph{Proceedings of the 54th Annual Meeting of the Association
  for Computational Linguistics (Volume 1: Long Papers)}, pages 588--598,
  Berlin, Germany. Association for Computational Linguistics.

\bibitem[{Song et~al.(2018)Song, Wang, Hamza, Zhang, and
  Gildea}]{Song2018LeveragingCI}
Linfeng Song, Zhiguo Wang, Wael Hamza, Yue Zhang, and Daniel Gildea. 2018.
\newblock \href {https://doi.org/10.18653/v1/N18-2090} {Leveraging context
  information for natural question generation}.
\newblock In \emph{Proceedings of the 2018 Conference of the North {A}merican
  Chapter of the Association for Computational Linguistics: Human Language
  Technologies, Volume 2 (Short Papers)}, pages 569--574, New Orleans,
  Louisiana. Association for Computational Linguistics.

\bibitem[{Sun et~al.(2018)Sun, Liu, Lyu, He, Ma, and
  Wang}]{Sun2018AnswerfocusedAP}
Xingwu Sun, Jing Liu, Yajuan Lyu, Wei He, Yanjun Ma, and Shi Wang. 2018.
\newblock \href {https://www.aclweb.org/anthology/D18-1427} {Answer-focused and
  position-aware neural question generation}.
\newblock In \emph{Proceedings of the 2018 Conference on Empirical Methods in
  Natural Language Processing}, pages 3930--3939, Brussels, Belgium.
  Association for Computational Linguistics.

\bibitem[{Sutskever et~al.(2014)Sutskever, Vinyals, and
  Le}]{Sutskever2014SequenceTS}
Ilya Sutskever, Oriol Vinyals, and Quoc~V Le. 2014.
\newblock \href
  {http://papers.nips.cc/paper/5346-sequence-to-sequence-learning-with-neural-networks.pdf}
  {Sequence to sequence learning with neural networks}.
\newblock In \emph{Advances in Neural Information Processing Systems 27}, pages
  3104--3112. Curran Associates, Inc.

\bibitem[{Vanderwende(2007)}]{Vanderwende2007AnsweringAQ}
Lucy Vanderwende. 2007.
\newblock Answering and questioning for machine reading.
\newblock In \emph{AAAI Spring Symposium: Machine Reading}.

\bibitem[{Wang et~al.(2019)Wang, Wei, Fan, Liu, and Huang}]{Wang2019AMC}
Siyuan Wang, Zhongyu Wei, Zhihao Fan, Yang Liu, and Xuanjing Huang. 2019.
\newblock A multi-agent communication framework for question-worthy phrase
  extraction and question generation.
\newblock In \emph{AAAI Conference on Artificial Intelligence}.

\bibitem[{Wang et~al.(2018)Wang, Liu, Huang, and Nie}]{Wang2018LearningTA}
Yansen Wang, Chenyi Liu, Minlie Huang, and Liqiang Nie. 2018.
\newblock \href {https://www.aclweb.org/anthology/P18-1204} {Learning to ask
  questions in open-domain conversational systems with typed decoders}.
\newblock In \emph{Proceedings of the 56th Annual Meeting of the Association
  for Computational Linguistics (Volume 1: Long Papers)}, pages 2193--2203,
  Melbourne, Australia. Association for Computational Linguistics.

\bibitem[{Yatskar(2018)}]{Yatskar2018AQC}
Mark Yatskar. 2018.
\newblock A qualitative comparison of coqa, squad 2.0 and quac.
\newblock \emph{CoRR}, abs/1809.10735.

\bibitem[{Yuan et~al.(2017)Yuan, Wang, Gulcehre, Sordoni, Bachman, Zhang,
  Subramanian, and Trischler}]{Yuan2017MachineCB}
Xingdi Yuan, Tong Wang, Caglar Gulcehre, Alessandro Sordoni, Philip Bachman,
  Saizheng Zhang, Sandeep Subramanian, and Adam Trischler. 2017.
\newblock \href {https://doi.org/10.18653/v1/W17-2603} {Machine comprehension
  by text-to-text neural question generation}.
\newblock In \emph{Proceedings of the 2nd Workshop on Representation Learning
  for {NLP}}, pages 15--25, Vancouver, Canada. Association for Computational
  Linguistics.

\bibitem[{Zhao et~al.(2018)Zhao, Ni, Ding, and Ke}]{Zhao2018ParagraphlevelNQ}
Yao Zhao, Xiaochuan Ni, Yuanyuan Ding, and Qifa Ke. 2018.
\newblock \href {https://www.aclweb.org/anthology/D18-1424} {Paragraph-level
  neural question generation with maxout pointer and gated self-attention
  networks}.
\newblock In \emph{Proceedings of the 2018 Conference on Empirical Methods in
  Natural Language Processing}, pages 3901--3910, Brussels, Belgium.
  Association for Computational Linguistics.

\bibitem[{Zhou et~al.(2017)Zhou, Yang, Wei, Tan, Bao, and
  Zhou}]{Zhou2017NeuralQG}
Qingyu Zhou, Nan Yang, Furu Wei, Chuanqi Tan, Hangbo Bao, and Ming Zhou. 2017.
\newblock Neural question generation from text: {A} preliminary study.
\newblock In \emph{Proceedings of the 6th {CCF} International Conference on
  Natural Language Processing and Chinese Computing ({NLPCC})}, pages 662--671,
  Dalian, China.

\end{thebibliography}
\bibliographystyle{acl_natbib}

\end{document}